\documentclass[letterpaper, 10 pt, conference]{ieeeconf}  

\IEEEoverridecommandlockouts                              

\usepackage[table]{xcolor}
\usepackage{graphicx}
\usepackage{booktabs}
\usepackage{multirow}
\usepackage{array}
\usepackage{siunitx}
\usepackage{adjustbox}
\usepackage{subcaption}
\usepackage{amssymb}
\usepackage{amsmath}
\usepackage[flushleft]{threeparttable}
\usepackage[compact]{titlesec}
\titlespacing{\section}{0pt}{*0.8}{*0.6}
\titlespacing{\subsection}{0pt}{*0.8}{*0.6}

\title{\LARGE \bf
BEV-ODOM: Reducing Scale Drift in Monocular Visual Odometry with BEV Representation
}

\author{Yufei Wei$^{1}$, Sha Lu$^{1}$, Fuzhang Han$^{1}$, Rong Xiong$^{1}$ and Yue Wang$^{1\dagger}$%
\thanks{*This work was supported by the National Key R\&D Program of China (Grant No. 2022YFB4701502), the National Nature Science Foundation of China (Grant No. 62373322), and the Zhejiang Provincial Natural Science Foundation of China (Grant No. LD24F030001), the "Leading Goose" R\&D Program of Zhejiang (Grant No. 2023C01177).}
\thanks{$^{1}$Laboratory of Industrial Control and Technology, and the Institute of Cyber-Systems and Control, Zhejiang University, Hangzhou, 310058, China.}
\thanks{$^{\dagger}$Corresponding author wangyue@iipc.zju.edu.cn.}
}

\begin{document}

\maketitle
\thispagestyle{empty}
\pagestyle{empty}

\begin{abstract}

Monocular visual odometry (MVO) is vital in autonomous navigation and robotics, providing a cost-effective and flexible motion tracking solution, but the inherent scale ambiguity in monocular setups often leads to cumulative errors over time. In this paper, we present BEV-ODOM, a novel MVO framework leveraging the Bird's Eye View (BEV) Representation to address scale drift. Unlike existing approaches, BEV-ODOM integrates a depth-based perspective-view (PV) to BEV encoder, a correlation feature extraction neck, and a CNN-MLP-based decoder, enabling it to estimate motion across three degrees of freedom without the need for depth supervision or complex optimization techniques. Our framework reduces scale drift in long-term sequences and achieves accurate motion estimation across various datasets, including NCLT, Oxford, and KITTI. The results indicate that BEV-ODOM outperforms current MVO methods, demonstrating reduced scale drift and higher accuracy. 

\end{abstract}

\section{Introduction}

Monocular visual odometry (MVO) has been of interest for years due to its cost-effectiveness, serving as a notable solution in robotics and autonomous driving. It acts as an affordable and easily deployable supplement to navigation aids like GPS and inertial navigation systems. Despite its advantages, MVO's widespread adoption is limited by a key challenge: scale ambiguity. Due to the lack of general depth information, monocular systems typically estimate motion on a relative scale.

Traditional MVO methods, such as feature-based methods \cite{mur2017orb, campos2021orb}, semi-direct methods \cite{forster2014svo}, and direct methods \cite{engel2017direct}, establish their scale during initialization, using it as a global reference. This approach closely links scale estimation with initial motion, making tracking performance highly sensitive to startup movement speed. Moreover, these methods heavily rely on the initial scale setting, resulting in a severe scale drift issue over time.

Learning-based MVO methods utilize the powerful fitting capabilities of machine learning to model the prior distributions in training data. \cite{konda2015learning, wang2017deepvo} use Convolutional Neural Networks (CNNs) to automatically extract features from images and regress poses based on temporal modeling methods. Additionally, methods such as \cite{zhao2021deep, zhang2021deep, wagstaff2020self, zhan2021df, bian2019unsupervised, li2020self, teed2021droid} combine the interpretability of traditional methods with the strong data-fitting abilities of deep learning. These methods incorporate deep learning into steps like absolute scale recovery and feature point selection to achieve absolute scale and enhance matching robustness. To achieve high depth estimation accuracy, these methods often introduce depth supervision or optical flow supervision as additional supervision, which brings additional costs.

\begin{figure}[t]
\centering
\includegraphics[width=0.46\textwidth]{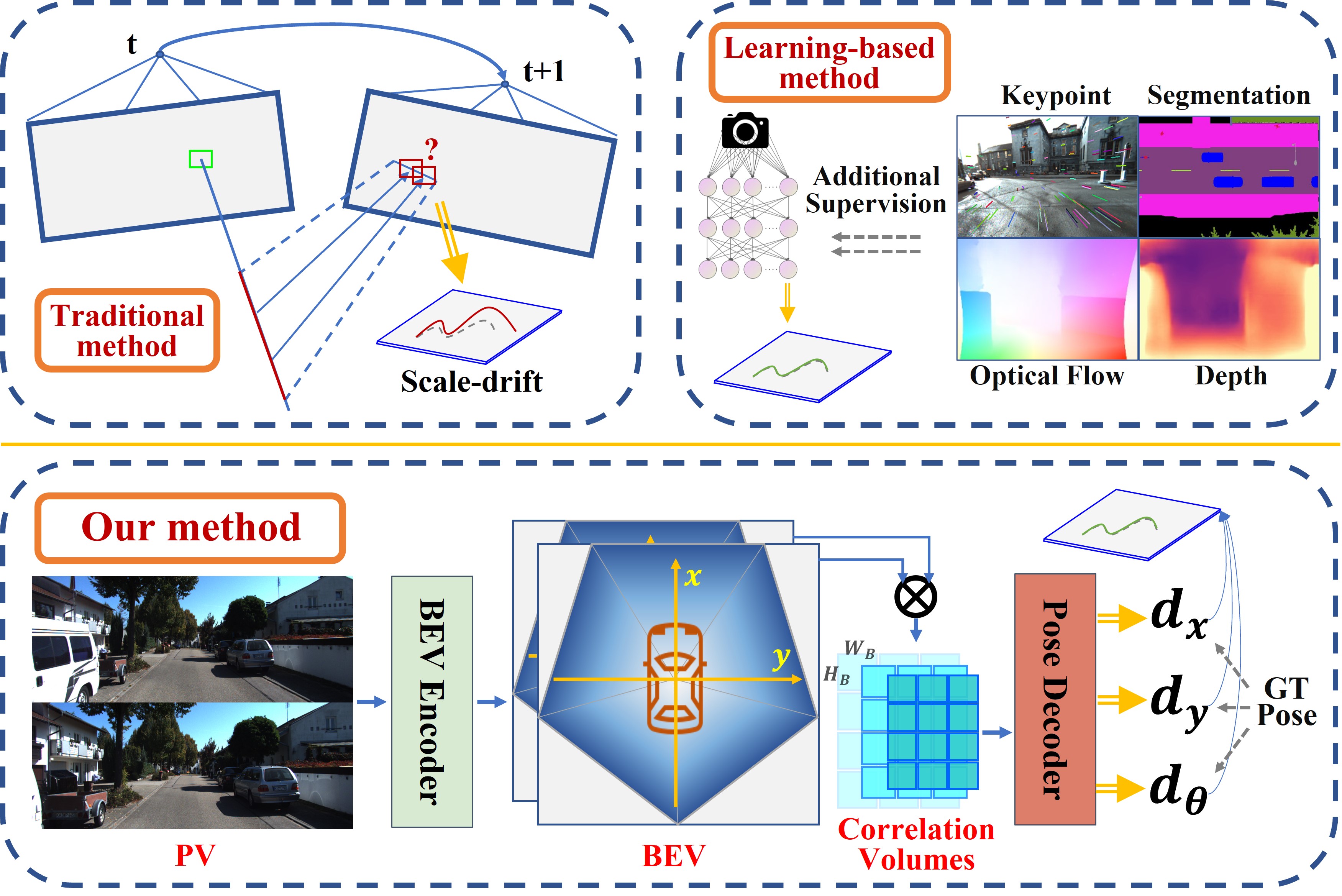}
\caption{Comparison of MVO approaches: traditional methods lack consistent scaling; learning-based methods require additional supervision. In contrast, our method achieves low scale drift using only pose supervision with BEV representation.}
\label{fig:pic1}
\vspace{-0.6cm}
\end{figure}

In recent years, with the advancement of BEV transformation techniques and the excellent performance of BEV representation in 3D detection and scene segmentation, some methods have begun to utilize BEV representation for visual odometry implementation, such as \cite{ross2022bev, zhang2022bev, li2023occ}. The motivation for using BEV representation lies in leveraging the common ground plane assumption in autonomous driving to simplify the six degrees of freedom (6-DoF) odometry estimation problem. However, these methods have not moved beyond the framework of other 3D tasks under BEV representation; they require scene segmentation first and then use the segmentation results to estimate the pose. The use of side-task supervision raises questions about whether the inherent scale properties come from the BEV representation itself or the supervision. Furthermore, these methods lead to high label acquisition costs and do not fully capitalize on the direct information provided by the BEV Representation.

To solve this problem, we propose BEV-ODOM, a novel approach to MVO utilizing a BEV representation. Our framework is structured around a depth-based perspective-view to Bird's-Eye-View encoder, a correlation feature extraction neck to assess the similarity between two BEVs with different shifts, and a decoder that integrates CNNs and Multi-Layer Perceptrons (MLPs) for estimating motion across three degrees of freedom (3-DoF). Different from existing learning-based MVO methods, our approach eschews complex procedures such as bundle adjustment, pose graph optimization, and side-task supervision. Different from other visual odometry methods that rely on BEV representation, our approach does not rely on segmentation results under BEV maps or occupancy maps for pose estimation, nor does it require additional supervision. This simplification not only enhances the efficiency of our method but also avoids the impact of inaccurate segmentation results on MVO and reduces data collection costs. By fully utilizing the consistent scale properties and the precise, detailed feature extraction capabilities of BEV representation, our method demonstrates excellent scale consistency and achieves state-of-the-art (SOTA) performance on challenging datasets under a 3-DoF evaluation. Because the NCLT and Oxford datasets exhibit minimal changes in z-axis translation, pitch, and roll, our method's performance remains equally excellent under a 6-DoF evaluation.

The contributions of our works are as follows:
\begin{itemize}

\item We propose a novel MVO framework utilizing BEV representation, effectively addressing scale drift and achieving better accuracy.
\item Our method simplifies the learning-based MVO pipeline from BEV representation, eliminating the need for supervision from side tasks including depth estimation, segmentation and occupancy map generation, improving its efficiency and robustness.
\item Our method achieves SOTA performance among current MVO methods on challenging datasets.

\end{itemize}

\section{Related Work}

\subsection{Traditional Methods}

Traditional MVO methods rely on geometric and feature-based approaches for motion estimation, which are robust in various environments without needing extensive training data. For example, ORB-SLAM2 \cite{mur2017orb} and ORB-SLAM3 \cite{campos2021orb} detect and describe keypoints using ORB features, then match these features across frames to estimate motion.
SVO \cite{forster2014svo} adopts a semi-direct approach, which uses both direct pixel intensity information and feature-based methods. It balances speed and accuracy, making it suitable for real-time applications.
DSO \cite{engel2017direct}, a direct method, focuses on minimizing the photometric error over a set of selected pixels across multiple frames, allowing for precise and efficient pose estimation.

In all these methods, scale is typically established through initialization and maintained using relative motion estimation. However, since monocular systems lack absolute depth information, they rely on scale consistency through frame-to-frame motion estimation and loop closures. ORB-SLAM methods, for instance, employ bundle adjustment and loop closure detection to correct scale drift and refine the map. SVO and DSO maintain scale by continuously optimizing the camera trajectory and the 3D map points.

\subsection{Learning-based Methods}

Learning-based methods initially use deep neural networks for direct pose regression. Recent approaches focus on providing absolute scale information for MVO by predicting monocular depth, optical flow, and other auxiliary tasks.

\cite{konda2015learning} utilizes a convolutional network for direct motion learning from images, demonstrating the capabilities of neural networks in visual odometry.
DeepVO \cite{wang2017deepvo} employs deep recurrent convolutional neural networks for end-to-end visual odometry, improving accuracy and robustness.

DDVO \cite{zhao2021deep}, DOC \cite{zhang2021deep}, and DPC \cite{wagstaff2020self} use deep learning-based pose correction to enhance pose estimation accuracy and scale consistency.

DF-VO \cite{zhan2021df} employs depth and bidirectional optical flow to filter feature points and uses a depth estimation network for precise 3D positioning, enabling comprehensive scale-inclusive pose estimation. In contrast, methods like \cite{bian2019unsupervised} and \cite{li2020self} utilize self-supervised learning for monocular depth prediction, which eliminates the need for extra depth supervision and partially reduces scale drift. However, as demonstrated in the experiments by DF-VO, these self-supervised methods suffer from scale ambiguity, resulting in lower performance compared to supervised methods like DF-VO.

DROID-SLAM \cite{teed2021droid} presents a novel method for motion estimation through feature extraction and inter-frame correlation, using a differentiable Dense Bundle Adjustment (DBA) layer for multi-frame optimization and camera pose refinement.

\begin{figure*}[t]
\centering
\includegraphics[width=\textwidth]{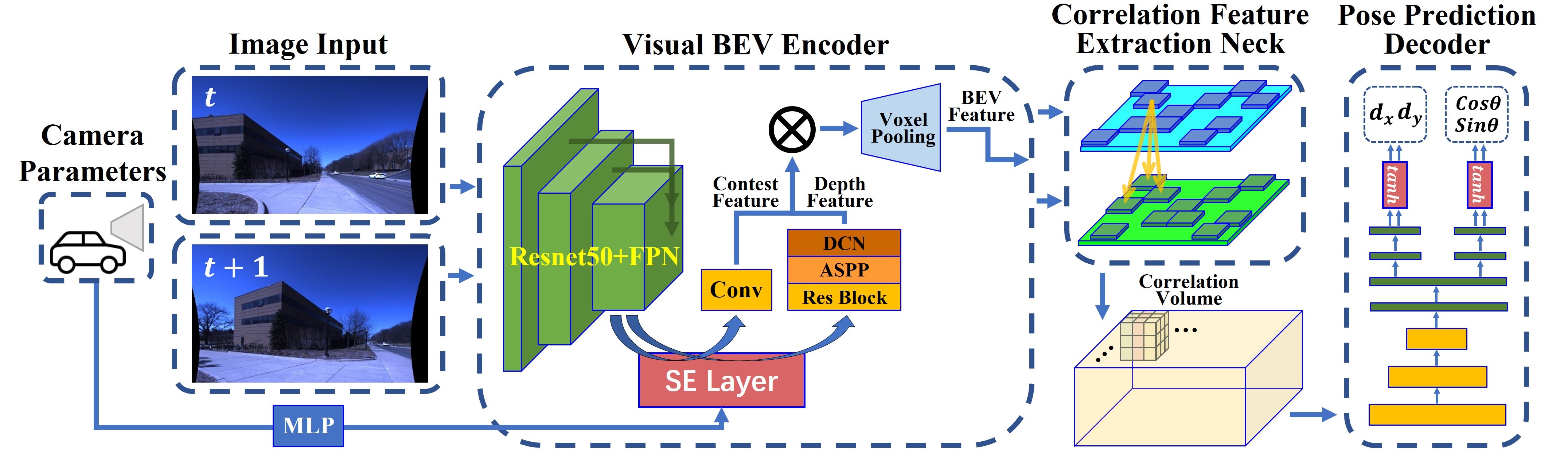}
\caption{Overview of the proposed framework.}
\label{fig:BEV-ODOM_net}
\vspace{-0.6cm}
\end{figure*}

\subsection{BEV Representation Methods}

In recent years, BEV-based odometry methods have gained much attention in autonomous driving and robot navigation. These methods use the ground plane assumption to simplify problems. BEV-SLAM \cite{ross2022bev} leverages deep learning networks to predict semantic segmentation maps under the BEV from monocular images and uses the enhanced correlation coefficient to find a suitable alignment in the BEV plane. BEV-Locator \cite{zhang2022bev} encodes surrounding images into BEV space and semantic map features into queries, employing a cross-model transformer to query the ego vehicle's localization through cross-attention. OCC-VO \cite{li2023occ} utilizes TPV-Former to transform multi-camera images into three-dimensional semantic occupancy point clouds, estimating each frame's pose through registration with a global semantic map. These methods extract key environmental features through supervised semantic segmentation to achieve motion estimation and map construction.

However, the performance of these methods strongly depends on accurate semantic information supervision, and obtaining high-quality semantic annotation data is both time-consuming and costly, especially in large-scale and complex environments. Furthermore, the robustness of these methods in the face of inaccurate or missing semantic information is a significant concern. Uncertainties in semantic information may stem from annotation errors, inaccurate model predictions, or dynamic changes in the environment, all of which can reduce the performance of the final VO and SLAM tasks.

The method proposed in this paper, BEV-ODOM, does not rely on any side-tasks. It directly extracts and utilizes BEV features from visual inputs to estimate motion. Our method uses only pose supervision, avoiding the need for external semantic information. This approach fully employs the positional features of BEV representation, demonstrating that the scale consistency and higher accuracy originate from the BEV representation itself.

\section{Method}
BEV-ODOM introduces an MVO method that leverages the inherent scale consistency of BEV representation for motion estimation. It requires no additional modules beyond visual input and pose supervision. The process starts with feature extraction from the PV, followed by mapping these features onto the BEV plane through frustum projection. Then we calculate correlations of multi-channel BEV feature maps between two frames at different shifts, identifying matches that reveal the ego-motion's translation and rotation. Finally, we use CNNs and MLPs to refine these features and generate the final output.

In the following sections, we will sequentially introduce the components of the system and detail their implementation processes. A system overview is presented in Fig. \ref{fig:BEV-ODOM_net}.

\subsection{Visual BEV Encoder}

We select the depth-based LSS \cite{philion2020lift} architecture for BEV construction in our work. This choice aligns with our goal to create consistently scaled maps, ensuring precise and stable feature localization within a BEV representation. Unlike BEVDepth \cite{li2023bevdepth}, our BEV encoder does not require additional depth supervision because our method is fully differentiable, allowing pose supervision to provide gradients for optimizing both the feature extraction and depth distribution prediction parts of the BEV encoder.

We use ResNet-50 \cite{he2016deep} as the backbone to extract image features from monocular input images. These features are then integrated across multiple scales using a Feature Pyramid Network (FPN) \cite{lin2017feature}, resulting in multi-scale PV image features with dimensions \(C \times H \times W\).

Subsequently, we use an MLP to encode the camera's intrinsic and extrinsic parameters, which are then element-wise multiplied with \(F_{IF}\) to produce the encoded PV feature map \(F_{PV}\). This Squeeze-and-Excitation (SE) operation \cite{hu2018squeeze} is represented as:
\begin{equation}
F_{PV} = \text{MLP}(E, I) \odot F_{IF},
\end{equation}
where \(E\) and \(I\) are the extrinsic and intrinsic camera parameters, and \(\odot\) denotes element-wise multiplication.

A convolutional network then processes the fused features to produce a feature map with dimensions $F_{PV_\text{C}} \in C \times H \times W$, where \(C\) represents the number of channels in the feature map. Using a similar process, we generate a depth distribution map with dimensions \(F_{PV_\text{D}} \in D \times H \times W\), where \(D\) represents the depth resolution.

It is important to note that the depth distribution map is not generated using depth supervision during training, but rather coupled with the correlation feature extraction neck and pose prediction decoder using pose supervision. Therefore, the predicted depth distribution does not need to correspond to the actual scale. Predicting depth distribution at the feature map level, rather than on the entire perspective view input image, simplifies the depth prediction process and increases tolerance to prediction errors by estimating the probabilities of different depths.

Next, we modify \(F_{PV_\text{C}}\) and \(F_{PV_\text{D}}\) to include singleton dimensions for alignment. The new dimensions are \(\mathbb{R}^{C \times 1 \times H \times W}\) for \(F'_{PV_\text{C}}\) and \(\mathbb{R}^{1 \times D \times H \times W}\) for \(F'_{PV_\text{D}}\).

Then, we perform an element-wise multiplication across the channel and depth distribution dimensions to obtain the multi-dimensional feature map:
\begin{equation}
F_{PV_\text{multi}} = F'_{PV_\text{C}} \odot F'_{PV_\text{D}},
\end{equation}
where \(F_{PV_\text{multi}}\), with dimensions \(\mathbb{R}^{C \times D \times H \times W}\), represents the activation of each pixel's features at various depths.

Finally, we define the spatial resolution of the BEV feature map and map the \(F_{PV_\text{multi}}\) to the BEV space using frustum projection. Within the BEV space, we employ efficient voxel pooling to compress information along the \(z\)-axis, producing a BEV feature map output with dimensions $F_{BEV} \in C_B \times X_\text{range} \times Y_\text{range}$, where \(C_B\) represents the number of channels in the BEV feature map.

\subsection{Correlation Feature Extraction Neck}

Based on the BEV features, we design a correlation calculation module to determine feature correspondences across adjacent frames by generating correlation volumes from two BEV feature maps. Different from RAFT \cite{teed2020raft}, which creates a 4D correlation volume by calculating dot products across all pixel pairs for global correlation, we focus on the local correlation within a limited range of displacements. This approach is specifically tailored to address the small displacements typically observed in BEV plane odometer applications, effectively reducing unnecessary computations while capturing essential movements.

Our module takes two BEV feature maps as input, each of size $F_{BEV} \in C_B \times X_\text{range} \times Y_\text{range}$. It then shifts the later frame's BEV feature map from \(-\Delta_x\) to \(\Delta_x\) in the \(x\)-direction and from \(-\Delta_y\) to \(\Delta_y\) in the \(y\)-direction. For each shift, the correlation score matrix \(C_s\), for each position \((x, y)\) is computed as:
\begin{equation}
C_s[x, y] = \sum_{c=1}^{C} F_{BEV1}[c, x, y] \cdot F_{BEV2}[c, x+\Delta_x, y+\Delta_y],
\end{equation}
where \(F_{BEV1}\) and \(F_{BEV2}\) are the BEV feature maps of consecutive frames, \(C\) is the number of channels in the BEV feature maps, and \(\Delta_x\) and \(\Delta_y\) represent the shift in the \(x\) and \(y\) directions on the BEV feature maps, respectively. The result \(C_s\) is a matrix of size \(X_\text{range} \times Y_\text{range}\), corresponding to the spatial dimensions of the BEV feature maps, capturing the correlation score at each spatial position. By considering every possible shift combination, we create a correlation volume with dimensions \(2\Delta_x\ \times 2\Delta_y\ \times X_\text{range} \times Y_\text{range}\). This 4D correlation volume provides a correlation score matrix for each possible displacement and details a feature space that captures the frames' relative motion.

\subsection{Pose Prediction Decoder}

In the pose prediction decoder, we first merge the \(2\Delta_x\) and \(2\Delta_y\) dimensions into a single dimension of \(4\Delta_x\Delta_y\), then use convolutional layers to decrease the dimensionality of the correlation volumes. Subsequently, we flatten the output and process it through fully connected layers, resulting in two branches: one for predicting \(x\) and \(y\) displacements, and the other for \(\cos\theta\) and \(\sin\theta\) predictions to circumvent the discontinuity issues of directly predicting \(\theta\), facilitating the network's learning of the correct mapping. After the output layer, we apply a \(\tanh\) function for post-processing, scaling the outputs to reasonable levels and reducing the impact of outliers. Our model employs the relative pose between two frames for supervision, calculating \(L_1loss\) for the rotation matrix and translation vector. These losses are then weighted and combined to obtain the final loss, which is used to update the network parameters.

The overall supervised loss \(L_{\text{Rt}}\) for pose prediction that only considers rotation around the z-axis and translations along the x and y axes can be described as follows:
\begin{equation}
L_{\text{Rt}} = \text{L1Loss}(t_{\text{pred}}, t_{\text{gt}}) + \alpha \cdot \text{L1Loss}(R_{\text{pred}}, R_{\text{gt}}),
\end{equation}
where \(t_{\text{pred}}\) and \(R_{\text{pred}}\) are the predicted translation vector and rotation matrix, respectively, and \(t_{\text{gt}}\) and \(R_{\text{gt}}\) are the ground truth translation vector and rotation matrix, respectively. The factor \(\alpha\) serves to balance the contributions of the translation and rotation errors to the total loss.

\renewcommand{\arraystretch}{1.0} 

\begin{table*}[t]
\centering
\begin{threeparttable}
\caption{PERFORMANCE ON NCLT, OXFORD, AND KITTI DATASETS}
\begin{tabular}{lcccccccc}
\toprule
& \multicolumn{4}{c}{NCLT} & \multicolumn{4}{c}{NCLT\_PART} \\
\cmidrule(lr){2-5} \cmidrule(lr){6-9}
& RTE(\%) & RRE(°/100m) & ATE(m) & ATE(m) & RTE(\%) & RRE(°/100m) & ATE(m) & ATE(m) \\
\midrule
ORB-SLAM3 \cite{campos2021orb} (w/ LC) & / & / & / & / & 44.3* & 48.56* & 36.65* & 29.04\(^{\dag}\) \\
ORB-SLAM3 \cite{campos2021orb} (w/o LC) & / & / & / & / & 45.00* & \underline{47.94*} & \underline{34.9*} & 26.88\(^{\dag}\) \\
DROID-SLAM \cite{teed2021droid} (w/o GBA) & 44.17* & \underline{10.67*} & \underline{245.05*} & \underline{127.79\(^{\dag}\)} & \underline{40.65*} & 53.71* & 46.32* & 45.02\(^{\dag}\) \\
DF-VO \cite{zhan2021df} (F. Model) & \underline{41.03*} & 25.52* & 414.64* & 175.19\(^{\dag}\) & 93.25* & 59.18* & 59.95* & \underline{25.30\(^{\dag}\)} \\
Ours (w/o DS) & \textbf{4.75**} & \textbf{2.08**} & \textbf{56.77**} & \textbf{57.60\(^{\dag}\)} & \textbf{9.07**} & \textbf{6.18**} & \textbf{9.35**} & \textbf{5.90\(^{\dag}\)} \\
\midrule
\midrule
& \multicolumn{4}{c}{Oxford} & \multicolumn{4}{c}{Oxford\_PART} \\
\cmidrule(lr){2-5} \cmidrule(lr){6-9}
& RTE(\%) & RRE(°/100m) & ATE(m) & ATE(m) & RTE(\%) & RRE(°/100m) & ATE(m) & ATE(m) \\
\midrule
ORB-SLAM3 \cite{campos2021orb} (w/ LC) & 254.23* & 16.26* & 1766.7* & 190.71\(^{\dag}\) & 194.87* & \underline{1.03*} & 834.24* & 73.4\(^{\dag}\) \\
ORB-SLAM3 \cite{campos2021orb} (w/o LC) & 952.41* & 10.81* & 5547.81* & 310.13\(^{\dag}\) & 282.08* & \textbf{1.00*} & 1189.59* & 93.04\(^{\dag}\) \\
DROID-SLAM \cite{teed2021droid} (w/o GBA) & 136.58* & \underline{1.43*} & 1184.86* & 174.33\(^{\dag}\) & 84.16* & 1.07* & 351.45* & 63.62\(^{\dag}\) \\
DF-VO \cite{zhan2021df} (F. Model) & \underline{28.26*} & 2.34* & \underline{158.55*} & \underline{86.13\(^{\dag}\)} & \underline{73.31*} & 8.91* & \underline{95.91*} & \underline{59.42\(^{\dag}\)} \\
Ours (w/o DS) & \textbf{6.54**} & \textbf{1.27**} & \textbf{93.77**} & \textbf{83.17\(^{\dag}\)} & \textbf{9.86**} & 2.22** & \textbf{38.89**} & \textbf{38.87\(^{\dag}\)} \\
\midrule
\midrule
& \multicolumn{4}{c}{KITTI seq.09} & \multicolumn{4}{c}{KITTI seq.10} \\
\cmidrule(lr){2-5} \cmidrule(lr){6-9}
& RTE(\%) & RRE(°/100m) & ATE(m) & ATE(m) & RTE(\%) & RRE(°/100m) & ATE(m) & ATE(m) \\
\midrule
ORB-SLAM3 \cite{campos2021orb} (w/ LC) & 3.31* & 0.38* & \underline{6.11*} & \underline{6.01\(^{\dag}\)} & 5.26* & \underline{0.25*} & 34.80* & 6.31\(^{\dag}\) \\
ORB-SLAM3 \cite{campos2021orb} (w/o LC) & 13.07* & \underline{0.24*} & 51.87* & 38.71\(^{\dag}\) & 7.53* & \underline{0.25*} & 16.28* & \underline{6.23\(^{\dag}\)} \\
DROID-SLAM \cite{teed2021droid} (w/o GBA) & 21.01* & 0.32* & 73.70* & 74.31\(^{\dag}\) & 18.73* & \textbf{0.23*} & 45.52* & 17.26\(^{\dag}\) \\
DF-VO \cite{zhan2021df} (Stereo Trained) & \underline{2.07**} & \textbf{0.23**} & 7.72** & 7.64\(^{\dag}\) & \textbf{2.06**} & 0.36** & \textbf{3.00**} & \textbf{2.73\(^{\dag}\)} \\
Ours (w/o DS) & \textbf{1.72**} & 0.39** & \textbf{6.35**} & \textbf{4.62\(^{\dag}\)} & \underline{3.61**} & 0.53** & \underline{8.42**} & 7.30\(^{\dag}\) \\
\bottomrule
\end{tabular}
\begin{tablenotes}[flushleft]
\item[] \textbf{w/ LC}: With loop closure optimization; \textbf{w/o LC}: Without loop closure optimization; \textbf{w/o GBA}: Without final global bundle adjustment optimization; \textbf{F. Model}: Use foundation model for bidirectional optical flow prediction and monocular depth estimation; \textbf{Stereo Trained}: Model trained using data collected from a stereo camera.
\item[*] Scaled by the first 10m's ground truth and aligned using $SE(3)$.
\item[**] Aligned using $SE(3)$.
\item[\({\dag}\)] Aligned using $Sim(3)$.
\end{tablenotes}
\label{table:main_table}
\end{threeparttable}
\vspace{-0.6cm}
\end{table*}

\renewcommand{\arraystretch}{1.0}

\section{Experiments}

\subsection{Implementation Details}

We evaluate our method using three datasets of varying difficulty: the University of Michigan North Campus Long-Term Vision and LIDAR Dataset (NCLT) \cite{carlevaris2016university}, the Oxford Radar RobotCar Dataset (Oxford) \cite{maddern20171}, and the KITTI-odometry Dataset (KITTI) \cite{geiger2013vision}. The NCLT dataset is the most challenging, with significant bumps and light intensity variations. The Oxford dataset contains more complex driving paths compared to KITTI, although both were collected using vehicles.

We compare our method with three algorithms: ORB-SLAM3, DF-VO, and DROID-SLAM. For the Oxford and NCLT datasets, we train on three sequences and test on one. For KITTI, we train on sequences 00-08 and test on sequences 09 and 10, following the standard evaluation protocol.

In terms of training details, we use a \(128 \times 128\) BEV scope centered around the vehicle with a resolution of \(0.8\)m, cropped to a \(32 \times 64\) feature map according to the monocular camera setup, and employ a \(7 \times 7\) correlation range for computations. We sample the training set at \(0-2\)m or \(0-3\)m intervals to simulate different speeds and frame rates, enhancing data diversity. Considering the prevalence of straight-line driving in the datasets, we also incorporate data featuring larger rotations to boost the model's performance. For the validation/testing sets, we test using a fixed frame distance to accumulate the final trajectory without any post-processing or optimization, as is common in MVO.

We use RTE, RRE, and ATE as evaluation metrics to measure the odometry trajectory's deviation from the ground truth. RTE measures the average translational RMSE drift over distances from \(100\) to \(800\)m, while RRE calculates the average rotational RMSE drift over the same distances. ATE measures the mean translation error between predicted camera poses and ground truth.

To ensure fairness in results, we employ two alignment methods. The first method adjusts the translation scale of the prediction to match the first 10 meters of ground truth, which is closer to real usage in MVO methods lacking a real-world scaling factor. Since DF-VO is trained on the KITTI dataset, we do not align its translation scale for testing on KITTI. Our method does not align the translation scale for any test dataset to highlight its ability to handle monocular scale drift. The second method uses $Sim(3)$ alignment for optimal performance alignment of different methods.

Since DF-VO's monocular depth estimation and bidirectional optical flow networks have not been trained on the NCLT and Oxford datasets, and these datasets do not provide sufficient ground truth for depth and optical flow training, we replace them with foundation models ZoeDepth \cite{bhat2023zoedepth} and Unimatch-Flow \cite{xu2023unifying} for the NCLT and Oxford datasets. To fairly compare odometry performance, we disable DROID-SLAM's final global bundle adjustment optimization and ORB-SLAM3's loop closure detection, focusing on real-time scenarios where future observations are unavailable.

\begin{figure}[t]
\centering
\includegraphics[width=0.46\textwidth]{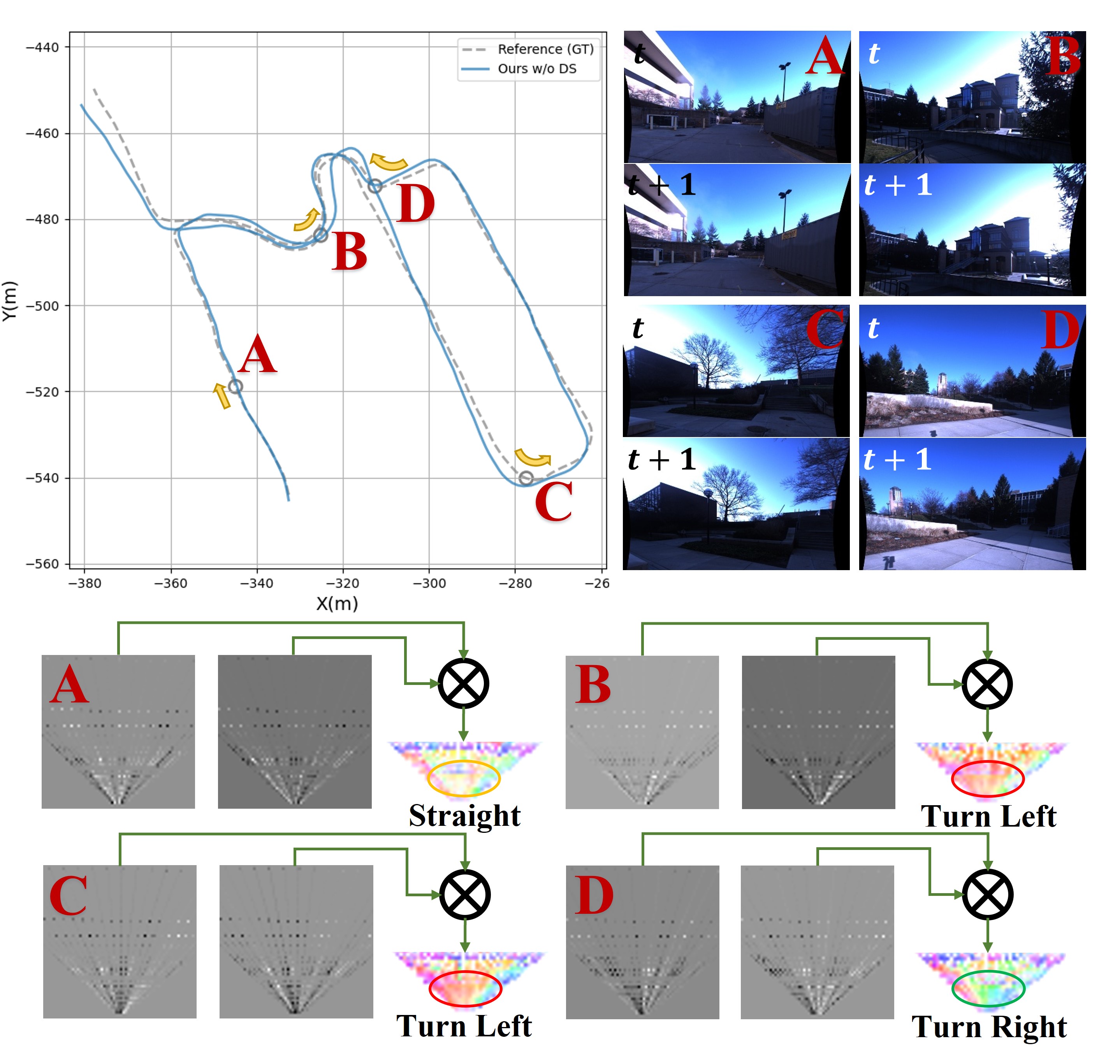}
\caption{BEV-ODOM's intermediate process and outcomes: predicted and actual trajectories (top left), camera images at four positions (A-D, top right), and the BEV feature maps' and BEV optical flow information's visualization (bottom).}
\label{fig:case_study}
\vspace{-0.6cm}
\end{figure}

\subsection{Case Study}

Fig. \ref{fig:case_study} visualizes the intermediate variables and outcomes of BEV-ODOM. The top left shows the predicted trajectory and the actual trajectory. The top right presents camera images from four example positions at times \(t\) and \(t+1\). In Fig. \ref{fig:case_study}, A shows a straight road segment, B and C show left turn scenarios, and D shows a right turn position. The bottom figure illustrates the feature maps encoded into BEV space by the visual BEV encoder and those with BEV optical flow information, extracted through the correlation feature extraction neck from these two frames.
It can be observed that during straight movement, left turns, and right turns, the feature maps with BEV optical flow information exhibit specific flow direction patterns. In contrast, predicting optical flow maps using perspective view input does not achieve similar effects due to varying environments and the distribution of near and far scenes. We infer that our Visual BEV Encoder can more effectively transform image features from the perspective view into BEV representation by predicting their depth distribution, thereby obtaining more stable BEV features. This results in more pronounced differences in BEV optical flow maps during various turns, allowing the MLPs to regress more stable and accurate results.

\begin{figure}[ht]
\centering

\begin{subfigure}{0.46\textwidth}
\centering
\includegraphics[width=\linewidth]{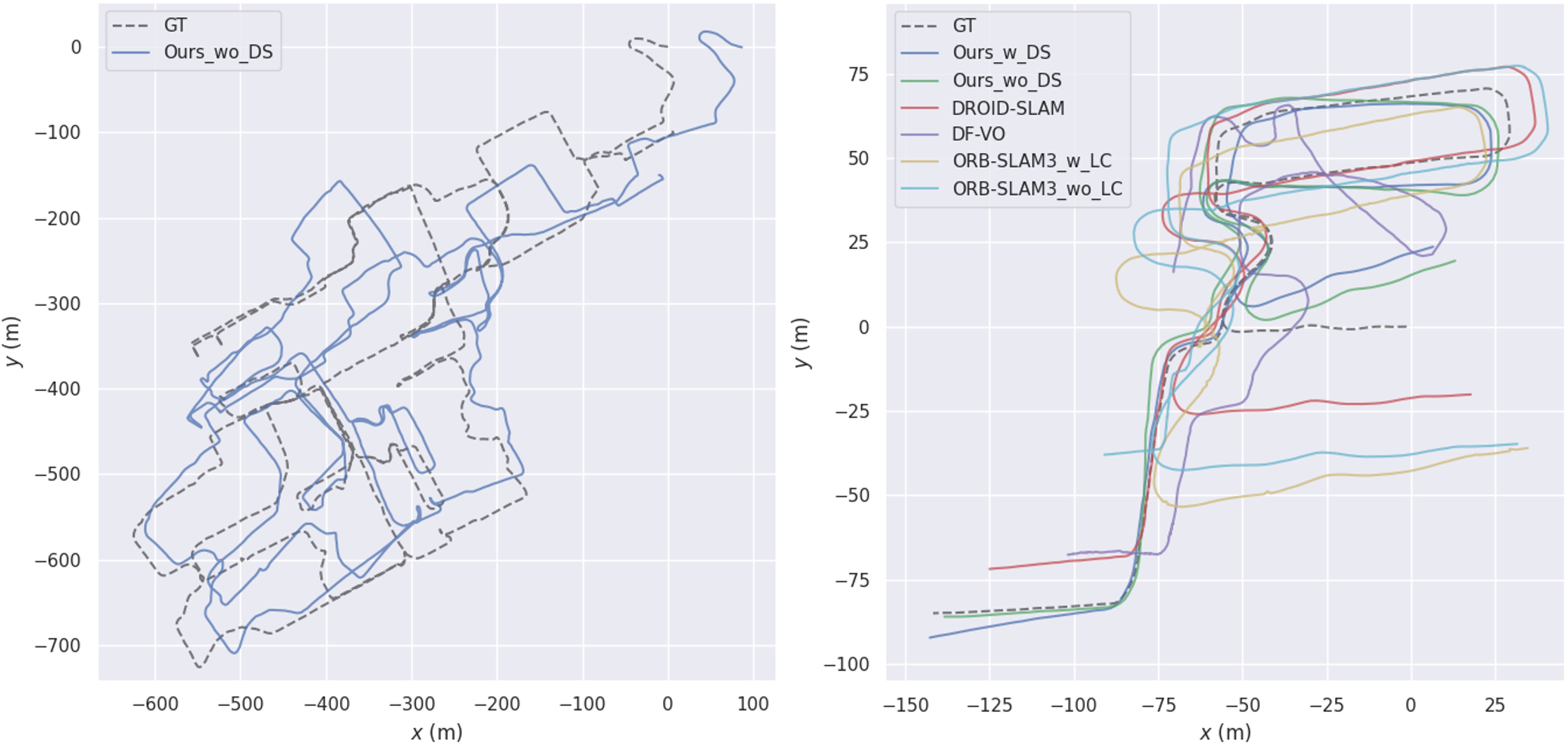}
\caption{Trajectories of our method and comparison methods on NCLT dataset: Left: full path test; Right: subset path test}
\end{subfigure}
\par\bigskip

\begin{subfigure}{0.46\textwidth}
\centering
\includegraphics[width=\linewidth]{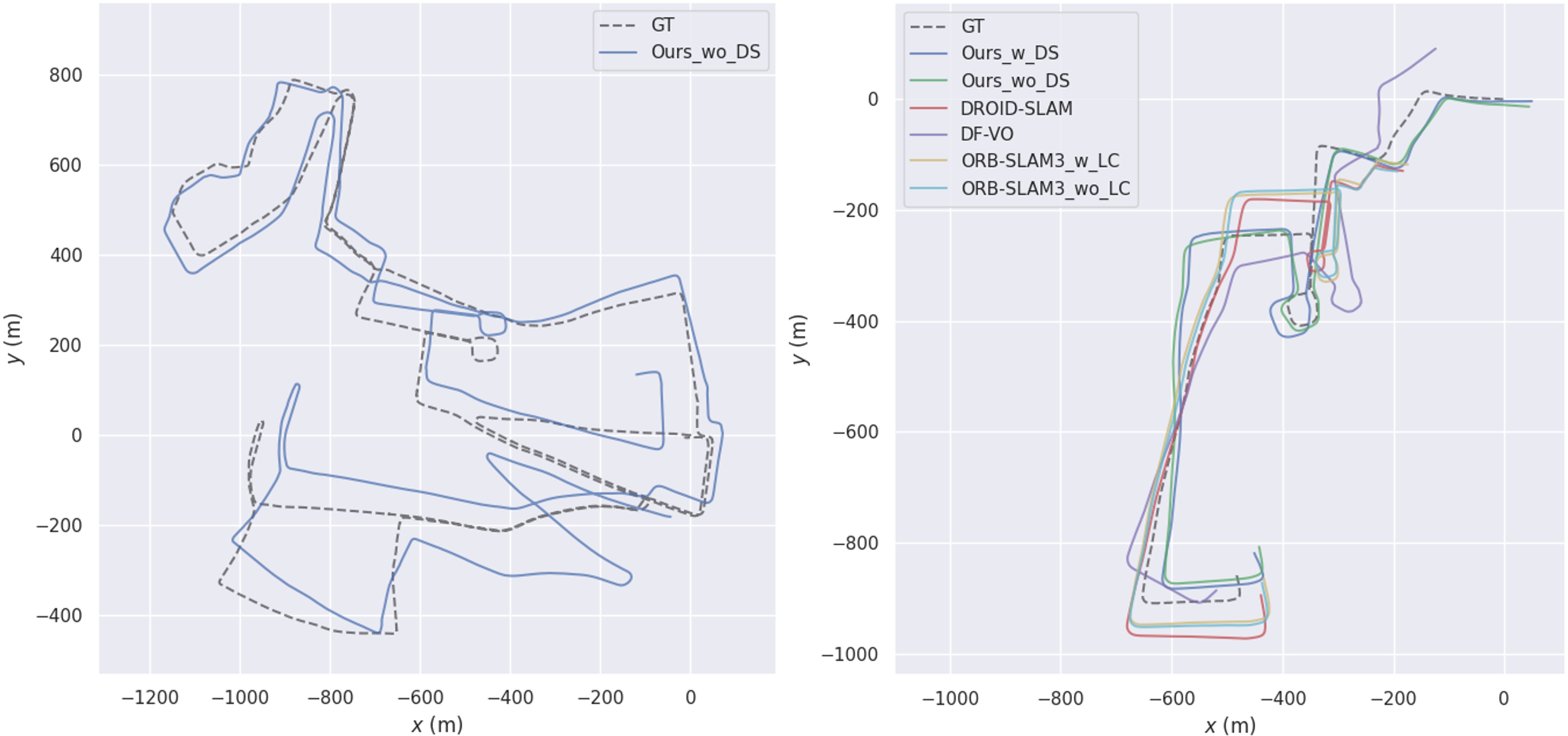}
\caption{Trajectories of our method and comparison methods on Oxford dataset: Left: full path test; Right: subset path test}
\end{subfigure}
\par\bigskip

\begin{subfigure}{0.46\textwidth}
\centering
\includegraphics[width=\linewidth]{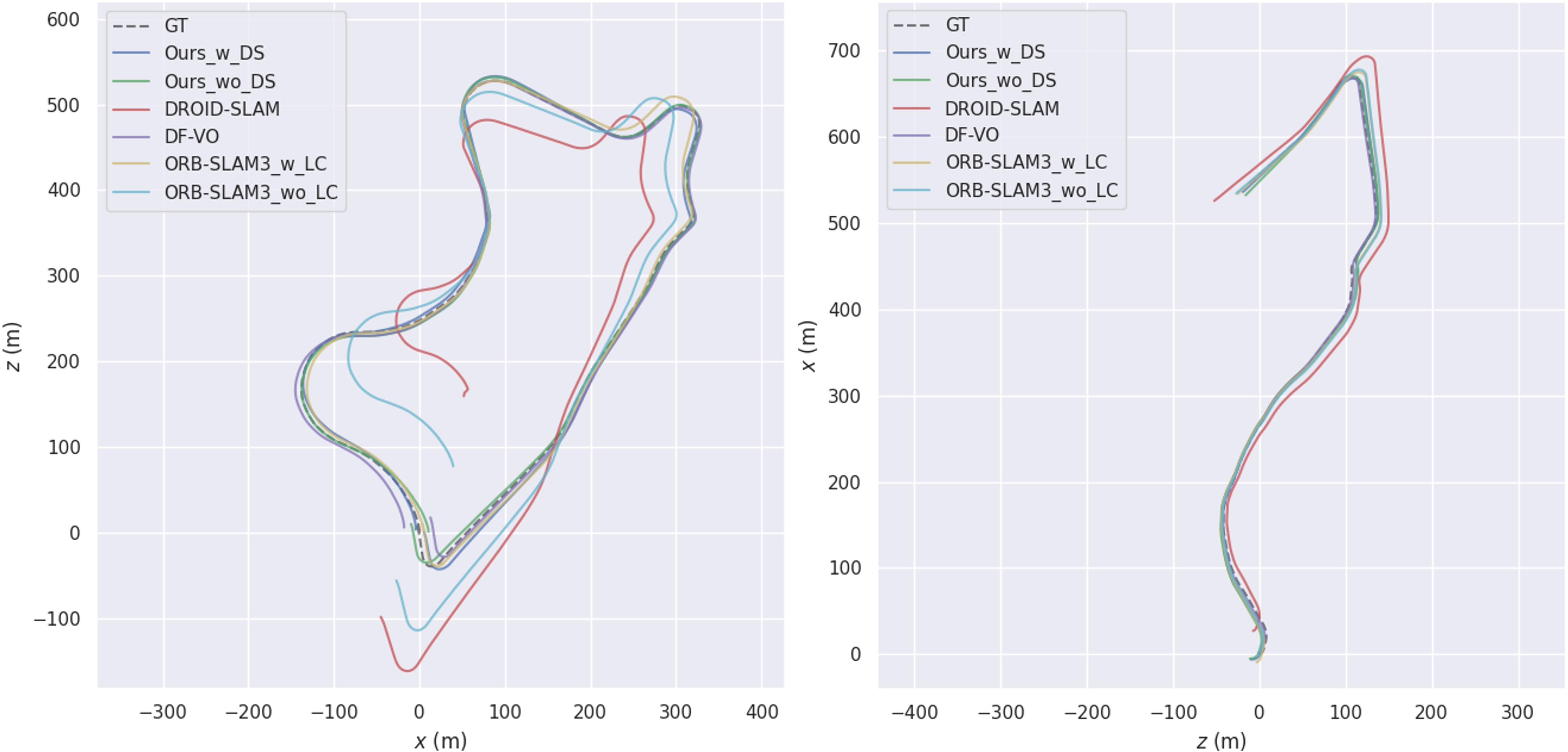}
\caption{Trajectories of our method and comparison methods on KITTI dataset: Left: seq.09; Right: seq.10}
\end{subfigure}

\caption{Trajectory comparisons on NCLT, Oxford, and KITTI datasets. For the NCLT (a) and Oxford (b) datasets, the left panels show full test paths and the right panels show selected subsets. For the KITTI (c) dataset, sequences 09 (left) and 10 (right) are displayed.}

\label{fig:trajectories_comparison}
\vspace{-0.4cm}
\end{figure}

\subsection{Trajectory Evaluation}

In the trajectory analysis, as shown in Fig. \ref{fig:trajectories_comparison}, for the NCLT and Oxford datasets, the left images depict the complete trajectories, while the right images show selected segments of these trajectories. This distinction is crucial for comprehensive evaluation:
\begin{itemize}
\item \textbf{Complete Trajectory Analysis (Left Images)}: These trajectory images are visualizations of results from training on three complete sequences and testing on another complete sequence. Only our method is shown because other methods often yield unreliable trajectories over the entire sequence. This highlights the challenges existing MVO methods face in maintaining scale consistency and accurate trajectory estimation over long durations or in complex environments.
\item \textbf{Selected Segment Analysis (Right Images)}: These trajectory images visualize results from training on part of one sequence and testing in completely unseen scenarios within the same sequence. This testing method evaluates our method's generalization ability and accuracy in new environments, with our method still performing the best in this evaluation.
\end{itemize}

This dual analysis demonstrates the robustness and adaptability of BEV-ODOM. It reliably handles long and complex trajectories and performs well in unfamiliar settings, highlighting the advantages of using BEV representation for MVO tasks in complex and long-distance scenarios that meet the flat plane assumption.

TABLE \ref{table:main_table} presents the performance comparison of our method against others on the KITTI, NCLT, and Oxford datasets.

In the KITTI dataset, our method achieves the best or second-best results on most metrics for seq.09, particularly excelling in Absolute Trajectory Error (ATE), which measures overall trajectory drift. On seq.10, our method's translation accuracy is not optimal, partly due to significant elevation changes throughout the sequence. This indicates room for improvement in our method for scenarios that do not meet the flat plane assumption. Another reason is that DF-VO uses stereo camera data for training.

For the more challenging NCLT and Oxford datasets, our method significantly outperforms others across almost all metrics.

DF-VO performs better on the KITTI dataset because it uses stereo depth information for training. However, even when tested on the NCLT and Oxford datasets with the most advanced foundation models providing bidirectional optical flow and monocular depth estimation, its performance remains suboptimal. This underscores the difficulties such methods face when the dataset lacks depth and flow supervision data or when the vehicle is not equipped with the necessary data collection devices. In contrast, our method achieves good scale consistency and precise relative pose estimation even without depth supervision. It excels in real-time scenarios without $Sim(3)$ alignment and achieves better ATE metrics on the NCLT and Oxford datasets compared to other methods that use $Sim(3)$ alignment.

Finally, our method demonstrates enhanced speed and lower memory consumption compared to techniques like DROID-SLAM, which requires continuous multi-frame optimization, and DF-VO, which involves predicting intermediate depth and bidirectional optical flow. Our approach achieves over \(60\) frames per second (fps) on an RTX4090 graphics card. This efficiency and reduced resource usage simplify the deployment of learning-based methods, making our approach more practical for real-world applications.

\subsection{Scale Drift Test}

TABLE \ref{tab:mean_scale_consistency} shows our method's outstanding performance in scale consistency. Firstly, we adjust the scale of the trajectories for all methods, except ours and DF-VO (on KITTI), according to the GT of the first 10 meters. Then, We calculate the scale drift using the following equation:
\begin{equation}
D_{\text{scale}} = \frac{1}{N} \sum_{i=1}^{N} \left| \log_2\left(\frac{d_i}{d_{i}^{GT}}\right) \right|,
\end{equation}
where \(D_{\text{scale}}\) is the average scale drift across all segments, \(N\) is the total number of segments, $d_i$ is the estimated displacement distance for the $i$-th segment, and $d_{i}^{GT}$ is the ground truth displacement distance for the $i$-th segment.

The reason for using the logarithm function and absolute value to measure scale drift is that the scale drift usually manifests not as a linear deviation but as a proportional difference from the ground truth. The logarithm function allows us to normalize these proportional differences, ensuring that scale overestimations and underestimations are equally penalized.

\renewcommand{\arraystretch}{1.0} 

\begin{table}[tbp]
\centering
\begin{threeparttable}
\caption{SCALE DRIFT ON THREE DATASETS}
\label{tab:mean_scale_consistency}
\setlength{\tabcolsep}{3pt}
\begin{tabular}{lcccc}
\toprule
Method & NCLT & Oxford & KITTI 09 & KITTI 10 \\
\midrule
ORB-SLAM3 \cite{campos2021orb} (w/o LC) & / & 1.2715* & 0.1281* & 0.1462* \\
ORB-SLAM3 \cite{campos2021orb} (w/ LC) & / & 0.1872* & 0.0316* & 0.2040* \\
DF-VO \cite{zhan2021df} & \underline{0.1821*} & \underline{0.2734*} & \underline{0.0297**} & \textbf{0.0260**} \\
DROID-SLAM \cite{teed2021droid} & 1.9033* & 0.5127* & 0.3406* & 0.3115* \\
Ours (w/o DS) & \textbf{0.0701**} & \textbf{0.1063**} & \textbf{0.0159**} & \underline{0.1042**} \\
\bottomrule
\end{tabular}
\begin{tablenotes}[flushleft]
\item[*] Scaled by the first 10m's ground truth and aligned using $SE(3)$.
\item[**] Aligned using $SE(3)$.
\end{tablenotes}
\end{threeparttable}
\vspace{-0.4cm}
\end{table}
\renewcommand{\arraystretch}{1.0}

\begin{figure}[tbp]
\centering
\includegraphics[width=0.46\textwidth]{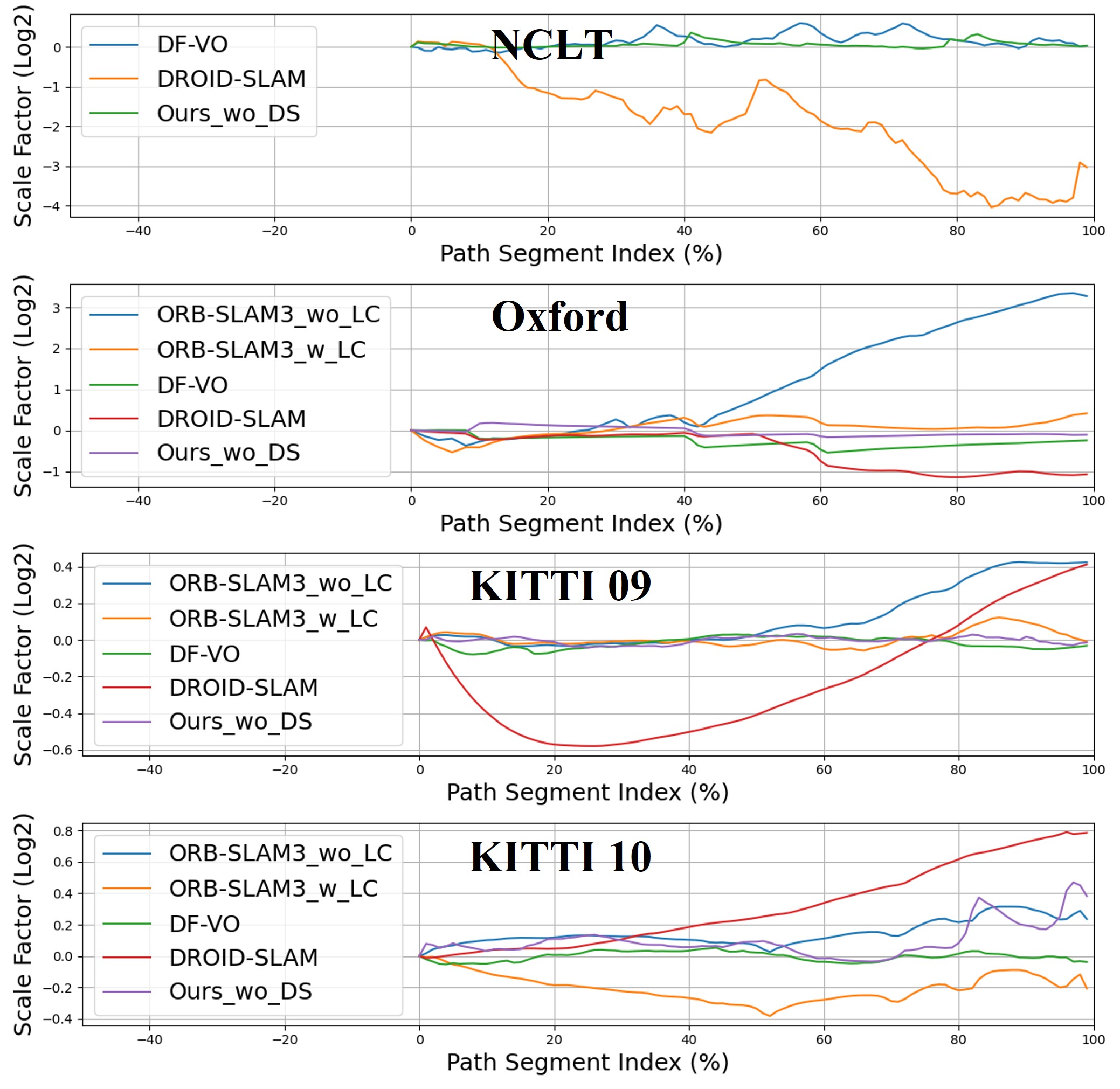}
\caption{Logarithmic scale factor variation along the path.}
\label{fig:SCALE}
\vspace{-0.5cm}
\end{figure}

We also present Figure \ref{fig:SCALE}, which shows the variation of logarithmic scale factors along the path for various methods. Compared to others, our method demonstrates a consistent scale factor throughout the entire path.

The results indicate that our method maintains a high level of performance across various datasets, only underperforming compared to the stereo-trained DF-VO method on one of the test sequences in the KITTI dataset. By observing Figure \ref{fig:SCALE} and analyzing the ground truth of this sequence, we find that the main cause of scale drift is the uphill slope in the last 50\% of the path. Since our method only predicts 3-DoF motion, this results in such error.

Additionally, it is worth noting that even in the severely shaking NCLT dataset, our method consistently maintains scale accuracy from start to finish. This occurs because BEV features exhibit less motion amplitude during severe shaking than PV features. The BEV encoder, which focuses on mapping features to the BEV space based on depth, is better equipped to manage these situations, whereas finding accurate matches for PV features is challenging during high-frequency, large-amplitude reciprocating movements.

\renewcommand{\arraystretch}{1.0} 

\begin{table*}[htbp]
\centering
\setlength{\tabcolsep}{4pt}
\begin{threeparttable}
\caption{ABLATION STUDY ON THREE DATASETS}
\begin{tabular}{
  >{\centering\arraybackslash}p{1.35cm} 
  >{\centering\arraybackslash}p{1.35cm} 
  >{\centering\arraybackslash}p{1.35cm} 
  >{\centering\arraybackslash}p{1.35cm} 
  >{\centering\arraybackslash}p{1.35cm} 
  cccccc
}
\toprule
\multirow{2}{*}{\parbox{1.35cm}{\centering BEV scope}} & \multirow{2}{*}{\parbox{1.35cm}{\centering BEV resolution}} & \multirow{2}{*}{\parbox{1.35cm}{\centering Feature cropped}} & \multirow{2}{*}{\parbox{1.35cm}{\centering Correlation range}} & \multirow{2}{*}{\parbox{1.35cm}{\centering Depth supervision}} & \multicolumn{2}{c}{NCLT} & \multicolumn{2}{c}{Oxford} & \multicolumn{2}{c}{KITTI} \\
\cline{6-11}
\noalign{\vskip 1mm}
&  &  &  &  & \multicolumn{1}{c}{RTE(\%)} & \multicolumn{1}{c}{RRE(°/100m)} & \multicolumn{1}{c}{RTE(\%)} & \multicolumn{1}{c}{RRE(°/100m)} & \multicolumn{1}{c}{RTE(\%)} & \multicolumn{1}{c}{RRE(°/100m)} \\
\midrule
128 & 0.8m & 32$\times$64 & 7$\times$7 & & \textbf{4.75} & 2.08 & 6.54 & 1.27 & \textbf{2.67} & \textbf{0.46} \\
128 & 0.8m & 32$\times$64 & 7$\times$7 & \checkmark & 5.27 & \textbf{1.94} & 6.61 & \textbf{1.10} & 3.22 & 0.57 \\
128 & 0.8m & \textbf{24$\times$64} & 7$\times$7 & & 4.77 & 2.22 & \underline{6.53} & \underline{1.23} & 3.90 & 0.60 \\
128 & 0.8m & \textbf{24$\times$64} & 7$\times$7 & \checkmark & 5.30 & \underline{1.96} & 7.07 & 1.51 & 3.66 & 0.72 \\
128 & 0.8m & \textbf{16$\times$64} & 7$\times$7 & & \underline{5.04} & 2.16 & \textbf{6.32} & 1.24 & \underline{2.93} & \underline{0.55} \\
128 & 0.8m & \textbf{16$\times$64} & 7$\times$7 & \checkmark & 6.16 & 2.85 & 7.17 & 1.68 & 3.78 & 0.67 \\
\textbf{256} & \textbf{0.2m} & 32$\times$64 & 7$\times$7 & & 11.14 & 5.12 & 8.07 & 1.79 & 8.90 & 2.47 \\
128 & 0.8m & 32$\times$64 & \textbf{9$\times$9} & & 5.40 & 2.21 & 6.73 & 1.39 & 3.04 & 0.64 \\
\bottomrule
\end{tabular}
\label{table:ablation_study}
\end{threeparttable}
\vspace{-0.4cm}
\end{table*}

\renewcommand{\arraystretch}{1.0}

\subsection{Ablation Study}

We design ablation experiments to analyze the performance of different hyperparameter combinations, as well as the network performance when introducing different supervisions and data. We employ RTE and RRE metrics as they provide insights into the odometry drift at various distance ranges. As shown in TABLE \ref{table:ablation_study}, we experiment with: 

\begin{itemize}

\item Different scope and precision of BEV feature maps, which affect the field of view range and the fineness of BEV.

\item Different feature map cropping sizes before correlation computation. Smaller maps would lose the positional information of surrounding features, while large maps would lead to higher computational costs and introduce irrelevant, out-of-view data.

\item Different shift distance ranges during correlation computation. This value affects the potential displacement range observed by the network. Setting this too large would also increase computational costs and introduce irrelevant information. We mainly chose this value based on the resolution of the BEV feature maps and the set frame sampling interval.

\end{itemize}

Through analyzing the results in the TABLE \ref{table:ablation_study}, we can conclude that the resolution of the BEV grid doesn't need to be very high, but expanding the overall coverage area ($scope \times resolution$) of the BEV feature map is beneficial. BEV feature maps that encompass a broader range of features not only provide implicit landmark information but also supply 'multi-scale' features to subsequent network layers. 'Multi-scale' refers to the varying granularity in the cross-correlation outputs between two BEV feature maps computed at different relative shifts. Due to BEV's resolution limits, nearby features on the BEV feature map yield coarse-scale but precisely located correlation outputs. Conversely, distant features display finer-scale correlation outputs with less precise locations due to larger displacements at the same rotation angle. 

The careful selection of feature map cropping sizes and shift distance ranges also impacts network performance. Experiments show that when using forward (rear) monocular image inputs, a $32 \times 64$ feature map crop and a $7 \times 7$ shift distance range are most suitable.

To validate our method's capability to achieve performance comparable to depth supervision without incorporating side task supervision, we conduct experiments using the same parameter settings. During training, we introduce depth supervision to the BEV encoder's depth prediction network and compare the final performance differences. The results show that our method's performance does not change significantly with or without depth supervision.

We speculate that this is because methods like BEVDepth, when introducing depth supervision, first downsample the depth map to match the ResNet's output dimensions. This operation guides the ResNet to uniformly focus on all points in the image when extracting features, rather than on landmark pixels that are more meaningful for BEV representation transformation. Additionally, depth data is usually collected by LiDAR, which can have issues such as synchronization discrepancies, extrinsic calibration errors, and sparse point clouds, leading to inaccurate depth supervision. In contrast, using pose supervision allows the BEV encoder to optimize the estimation of feature depth distribution through backpropagation, without limiting its ability to capture more effective key points from perspective views.

\section{Conclusion}
In this paper, we propose BEV-ODOM, a visual odometry framework designed to address the scale drift problem in visual odometry systems. We extract features from perspective view images and estimate their depth distribution, projecting them into 3D space and compressing them into BEV representation. Next, we use a correlation feature extraction module to capture motion information between the BEV feature maps. Finally, a CNN-MLP-based pose decoder estimates the 3-DoF motion. We conduct extensive experiments on the widely used NCLT, Oxford, and KITTI datasets to verify the effectiveness of our method. The results show that the proposed approach achieves superior performance across all datasets.

\addtolength{\textheight}{-12cm}   










\small
\bibliographystyle{ieeetr}
\bibliography{root}

\end{document}